\documentclass[]{mosi}

\usepackage{helvet}

\usepackage{amsmath} 
\usepackage{natbib}
\usepackage{graphicx}
\usepackage{subcaption} 

\usepackage[toc,page,header]{appendix}
\usepackage[utf8]{inputenc} 
\usepackage[T1]{fontenc}    
\usepackage{hyperref}
\usepackage{url}            
\usepackage{booktabs}       

\usepackage{amsfonts}       
\usepackage{nicefrac}       
\usepackage{microtype}      
\usepackage{wrapfig}
\usepackage{amssymb}        
\usepackage{titletoc}
\usepackage{minitoc}

\usepackage{array}
\usepackage{etoolbox}

\definecolor{lightblue}{RGB}{200, 230, 255}  
\definecolor{headerblue}{RGB}{150, 200, 255} 

\usepackage{pgfplots}
\usepackage{xcolor}
\usepackage{float}
\usepackage{comment}
\usepackage{multirow}
\usepackage{makecell}
\usepackage{siunitx}
\usepackage{tikz}
\usepackage{pgf-pie}
\usepackage{algpseudocode}
\usepackage[export]{adjustbox}
\usepackage{ragged2e}
\usepackage{tabularx}
\usepackage{caption}
\usepackage{enumitem}
\usepackage{pifont}
\usepackage[hang,flushmargin]{footmisc}
\usepackage{tcolorbox}
\tcbuselibrary{breakable}
\tcbuselibrary{skins}
\usepackage{listings}
\usepackage{threeparttable}
\usepackage{setspace}
\usepackage{needspace}
\usepackage{placeins}

\newcounter{methodalgorithm}
\renewcommand{\themethodalgorithm}{\arabic{methodalgorithm}}

\usepackage[scheme=plain,fontset=none]{ctex}

\usepackage[table]{xcolor}

\definecolor{oursgray}{gray}{0.95}
\definecolor{h3colblue}{HTML}{EAF2FF}

\definecolor{MossCyan}{HTML}{82D9FF} 
\definecolor{MossBlue}{HTML}{82B1FF}

\definecolor{tickG}{HTML}{00C853}
\definecolor{crossR}{HTML}{FF1744}

\newcommand{\hflogo}{\raisebox{-0.12em}{\includegraphics[height=1em]{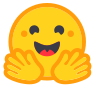}}}
\newcommand{\faGithub}{\raisebox{-0.12em}{\includegraphics[height=1em]{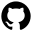}}}

\newtcolorbox{promptbox}[2][]{
    colback=white,
    coltext=black,
    arc=3mm,
    boxrule=0.5pt,
    colframe=black!60!white,
    title={#2},
    colbacktitle=black,
    coltitle=white,
    fonttitle=\bfseries,
    top=8pt,
    bottom=8pt,
    left=10pt,
    right=10pt,
    breakable,
    before upper={%
        \linespread{1}\selectfont
        \setlength{\parskip}{1ex plus 0.2ex minus 0.2ex}%
        \setlength{\parindent}{0pt}%
    },
    #1
}

\title{ABOPD: Antibody CDR Design via On-Policy Distillation}

\author{
Zhuo Yang$^{1,6}$,
Jiaying He$^{1,2}$,
Jiaqing Xie$^{3}$,
Daolang Wang$^{5}$,
Xipeng Qiu$^{1,3}$,
Yuxin Wang$^{1}$,
Tianfan Fu$^{7,4,\dagger}$,
Beilun Wang$^{6,\dagger}$
\\[2mm]
{\normalfont \normalsize $^{1}$Shanghai Innovation Institute}\\
{\normalfont \normalsize $^{2}$Zhejiang University}\\
{\normalfont \normalsize $^{3}$Fudan University}\\
{\normalfont \normalsize $^{4}$Shanghai Artificial Intelligence Lab}\\
{\normalfont \normalsize $^{5}$Xiamen University}\\
{\normalfont \normalsize $^{6}$Southeast University}\\
{\normalfont \normalsize $^{7}$Nanjing University}\\
{\normalfont \normalsize $^{\dagger}$Corresponding author.}
}

\abstract{
Antibodies are essential therapeutic molecules, and their complementarity-determining regions (CDRs) form the primary antigen-recognition interface.
Recent protein generative models have demonstrated broad capabilities in biomolecular design, yet post-training strategies for downstream objectives remain limited.
Standard denoising training operates on noisy states obtained by perturbing native structures, whereas recursive generation proceeds through model-generated intermediate states.
For flexible antibody CDR loops such as CDR-H3, this mismatch can allow backbone deviations to accumulate along the denoising trajectory and compromise antigen-facing loop geometry.
We introduce ABOPD, an \underline{A}nti\underline{b}ody design framework built on \underline{O}n-\underline{P}olicy \underline{D}istillation, leveraging privileged native geometry during training to supervise states visited along the model's own denoising trajectories.
With this fine-grained structural supervision, ABOPD substantially improves structural recovery on RAbD CDR-H3 generation, reducing RMSD by 0.42~\AA{} (2.37$\rightarrow$1.95~\AA{}) and outperforming supervised fine-tuning and offline distillation controls, offering a path to higher-fidelity protein design.

}

\checkdata[\faGithub\ GitHub]{\url{https://github.com/little1d/ABOPD}}
\checkdata[\hflogo\ Hugging Face]{\url{https://huggingface.co/collections/little1d/ABOPD}}

\begin{document}
\maketitle

\section{Introduction}
\begin{figure}[t]
\centering
\includegraphics[width=0.8\linewidth]{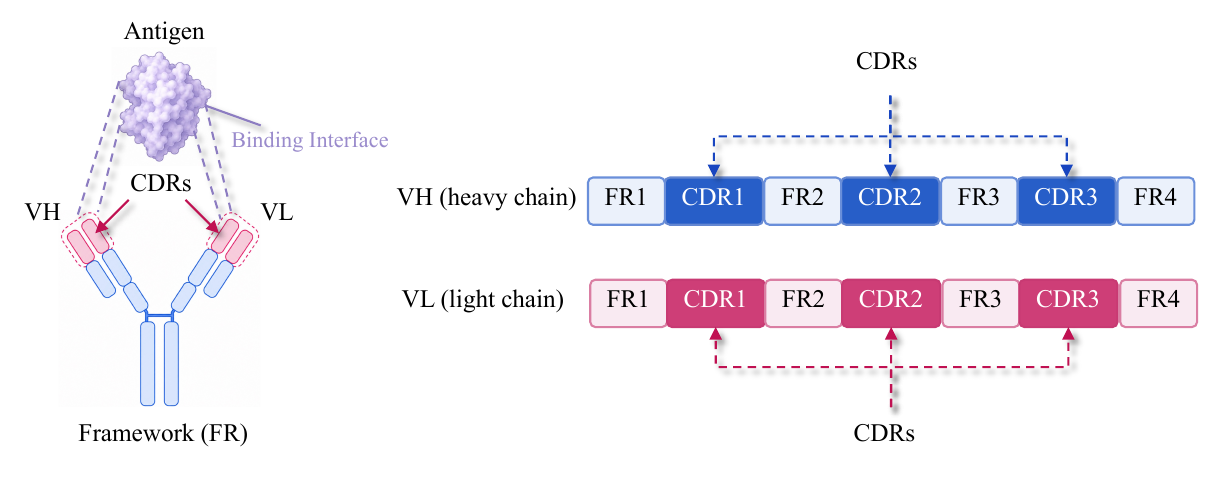}
\caption{\textbf{Antibody CDR design.} The left panel locates the six CDR loops at the antigen-binding interface, while the right panel shows their organization within the variable domains of the heavy and light chains.}
\label{fig:cdr-design-task}
\end{figure}

Computational design of antibody complementarity-determining regions (CDRs) for antigen-specific recognition is a central goal in protein engineering. This task remains challenging because antigen recognition depends critically on the coupling between the amino-acid sequences and 3D conformations of these flexible loops. The six CDRs (H1--H3 on the heavy chain and L1--L3 on the light chain) form most of the antigen-contacting surface \citep{pantazes2010optcdr,adolf2018rabd}. Consequently, CDR design requires jointly generating amino-acid sequences and loop conformations while preserving framework anchoring and geometric compatibility with the antigen interface (Figure~\ref{fig:cdr-design-task}). Learning the joint distribution of CDR sequences and structures is further complicated by the limited availability of experimentally determined antibody--antigen complex structures~\citep{schneider2022sabdab}.

Recent work has advanced antigen-conditioned CDR generation by leveraging geometric representations, equivariant architectures, and diffusion-based generative modeling \citep{luo2022diffab,kong2023dymean,lin2024geoab,martinkus2023abdiffuser}. Despite these advances, reliably generating CDRs with accurate loop geometry remains challenging. Existing work has largely emphasized model architectures and structural representations, while post-training strategies for antibody CDR generation remain comparatively underexplored. Post-training therefore offers a complementary route to improve existing antibody generators and, more broadly, adapt protein generative models to downstream design objectives, such as specialization to particular protein families \citep{madani2023large} and property optimization \citep{zhou2024depo}.

For diffusion-based CDR models, a key post-training challenge is the mismatch between the state distribution used for supervision and the one encountered during generation \citep{ning2023input}. Standard denoising training constructs noisy inputs by perturbing native CDR sequences and structures, which we call reference-derived states. Supervised fine-tuning (SFT) retains the denoising objective on this distribution, while offline distillation supervises the student with teacher predictions on the same class of states. During generation, however, reverse denoising is recursive: each step conditions on states produced by earlier model predictions. Prediction errors can therefore move subsequent states away from the reference-derived distribution, so supervision confined to this distribution does not directly address errors arising along the model's own generation trajectory. This mismatch motivates augmenting the denoising objective with supervision on states visited along the student's own reverse trajectories.

Such a mismatch is especially consequential for antibody CDR generation because flexible loops must satisfy geometric constraints. CDR-H3 exemplifies this challenge: it is typically the most diverse and conformationally flexible CDR, yet it must remain anchored to the antibody framework at both ends and adopt a conformation compatible with the antigen interface \citep{adolf2018rabd,luo2022diffab}. A C$\alpha$ backbone error introduced during reverse denoising can alter the loop's placement or antigen-facing orientation. Subsequent steps then condition on the resulting geometry, allowing the error to propagate and increasing the risk of steric clashes. Because residue identities and orientations are predicted within the same evolving geometric context, backbone drift may also influence subsequent sequence and orientation predictions. Thus, the state-distribution mismatch can manifest as accumulated geometric error in CDR backbone recovery.

To address this mismatch, we introduce ABOPD, an on-policy distillation
framework for post-training diffusion-based antibody CDR models. ABOPD adapts
a backbone-aware teacher using native geometry available during training and
distills C$\alpha$ coordinate-transition targets on states visited along the
student's reverse trajectories. The original denoising objective is retained
as an offline anchor for sequence, coordinate, and orientation prediction.
Our main contributions are:
\begin{itemize}[leftmargin=*]
    \item We identify a state-distribution mismatch in CDR diffusion that widens
    during denoising, while teacher corrections become most useful during later
    refinement.
    
    \item We propose ABOPD, combining on-policy C$\alpha$ coordinate-transition
    distillation on student rollout states with an offline denoising anchor.
    
    \item We demonstrate improved backbone recovery on RAbD CDR-H3 and simultaneous
    multi-CDR redesign. On RAbD, ABOPD reduces CDR-H3 RMSD from 2.37 to
    1.95~\AA{} and outperforms SFT and offline distillation.
\end{itemize}

\section{Related Work}
\subsection{Antibody CDR Design}

Antibody CDR design spans sequence modeling, energy-guided loop optimization,
and antigen-conditioned sequence--structure generation. Sequence models learn
antibody repertoire distributions or infill variable regions, while classical
pipelines search CDR sequences and conformations through loop sampling,
biophysical scoring, and Rosetta-based optimization
\citep{shuai2023iglm,pantazes2010optcdr,lapidoth2015abdesign,adolf2018rabd}.
Neural co-design methods couple residue identities with 3D geometry to generate
CDR sequences and structures using contextual information from the antibody
and antigen
\citep{kong2022mean,jin2022iterative,kong2023dymean,lin2024geoab,martinkus2023abdiffuser}.
Recent antibody and protein-binder models further incorporate all-atom
representations, flow-based generation, and interface-aware conditioning
\citep{watson2025rfantibody,iggm2025,wang2026abflow,zhang2025odesign,ppiflow2026}.
ABOPD complements this line of work by post-training an existing CDR diffusion
model \citep{luo2022diffab}.

\subsection{Structure-Aware Protein Generative Models}

More broadly, structure-aware protein generation spans sequence design under
structural constraints and direct generation of 3D protein geometry.
Hallucination-based methods optimize sequences through learned structure
predictors \citep{anishchenko2021hallucination}, while inverse-folding models
generate sequences conditioned on fixed backbones
\citep{dauparas2022proteinmpnn}. Diffusion and flow-based models further enable
direct backbone generation through coordinate and rigid-frame representations,
including diffusion and flow trajectories on $\mathrm{SE}(3)$
\citep{watson2023rfdiffusion,ingraham2023chroma,yim2023framediff,
yim2023frameflow,bose2024foldflow}. These geometric formulations provide the
methodological foundation for modern antibody CDR generation.

\subsection{Post-Training Strategies for Generative Models}

Post-training strategies differ in both their feedback signals and the states
on which those signals are applied. On-policy distillation applies teacher
feedback to student-generated states, reducing distribution mismatch in
autoregressive language models \citep{agarwal2024gkd}; recent diffusion and
flow methods extend this principle to continuous generative trajectories
\citep{jiang2026dopsd,fang2026flowopd,li2026diffusionopd}.
Reinforcement learning and preference optimization instead adapt generators
through task-specific rewards or preferences, with applications to molecular
design
\citep{olivecrona2017molecular,you2018gcpn,zhou2018moldqn,yang2025molact}
and protein design
\citep{subramanian2024proteinrl,ektefaie2024rldif,zhou2024depo}.
Related but distinct approaches include conditional guidance, which steers
sampling at inference \citep{ho2022cfg}, and privileged learning, which uses
information available only during training
\citep{vapnik2015learning,lopezpaz2016unifying}.
ABOPD combines on-policy distillation with training-time privileged structural
supervision for antibody CDR generation.

\section{Methods}
\label{sec:methods}
\begin{figure}[t]
\centering
\includegraphics[width=0.9\linewidth]{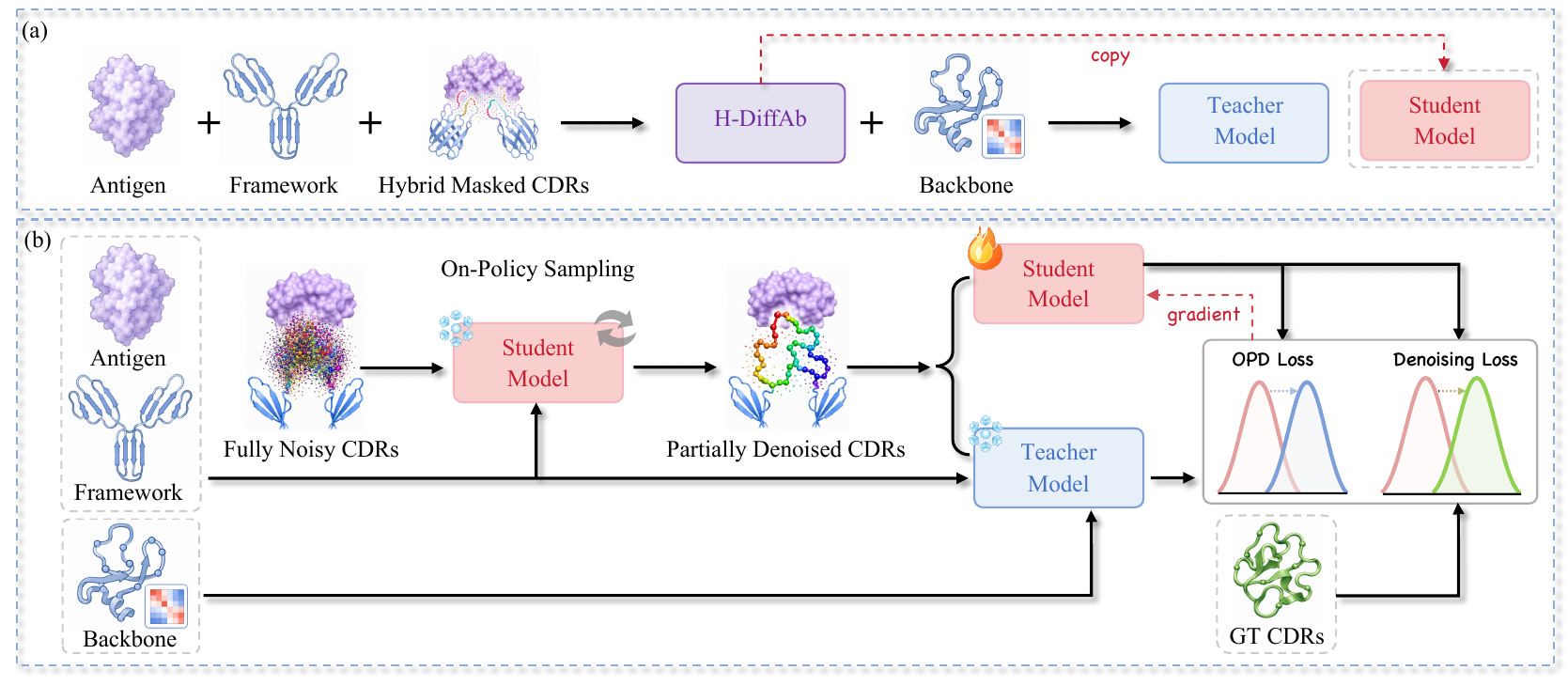}
\caption{\textbf{Overview of ABOPD.}
\textbf{(a)} Hybrid pretraining produces H-DiffAb, which initializes the student
and is adapted into a backbone-aware teacher using privileged native geometry.
\textbf{(b)} The frozen teacher supplies C$\alpha$ coordinate-transition targets
on student rollout states, while the offline denoising anchor preserves the
original training objective.}
\label{fig:framework}
\end{figure}

ABOPD aligns structural supervision with states encountered during recursive
generation while retaining the original denoising objective as an offline
anchor. As shown in Figure~\ref{fig:framework}, the framework comprises three
stages: hybrid pretraining to obtain H-DiffAb, backbone-aware teacher adaptation
using privileged native geometry, and student post-training via on-policy
C$\alpha$ coordinate-transition distillation.

\subsection{Problem Setup}

We study antigen-conditioned generation of one or more antibody CDRs. Let $\mathcal{G}$ be the set of target residues. Each target residue $i\in\mathcal{G}$ is represented by
\[
r_i=(s_i,x_i,O_i),
\]
where $s_i$ is its amino-acid identity, $x_i\in\mathbb{R}^3$ is its C$\alpha$ coordinate, and $O_i\in\mathrm{SO}(3)$ is its local orientation. The context $C$ contains the antigen and all non-target antibody residues, including the framework and any CDRs not selected for generation. Given $R_{\mathcal{G}}=\{r_i:i\in\mathcal{G}\}$, the model represents
\[
p_\theta(R_{\mathcal{G}}\mid C).
\]

Let $R_0$ denote the native target sequence and structure. Following DiffAb \citep{luo2022diffab}, the forward process independently perturbs amino-acid identities, C$\alpha$ coordinates, and residue orientations using categorical, Gaussian, and $\mathrm{SO}(3)$ diffusion, respectively. We use
\[
R_t=(s_t,x_t,O_t)\sim q_t(R_t\mid R_0,\mathcal{G})
\]

to denote a \emph{reference-derived state}, i.e., a noisy state obtained by perturbing the native CDR. Given $(R_t,C,t)$, the denoiser predicts a categorical reverse distribution,
coordinate noise, and a denoised orientation:
\[
f_\theta(R_t,C,t)
=
\left(
p_\theta^{\mathrm{seq}},
\hat{\epsilon}_\theta^{\mathrm{pos}},
\hat{O}_\theta^{t-1}
\right).
\]
Following \citet{luo2022diffab}, the standard supervised objective is
\begin{equation}
\label{eq:denoise}
\mathcal{L}_{\mathrm{denoise}}
=
\mathcal{L}_{\mathrm{seq}}
+
\lambda_{\mathrm{pos}}\mathcal{L}_{\mathrm{pos}}
+
\lambda_{\mathrm{ori}}\mathcal{L}_{\mathrm{ori}},
\end{equation}

where the three branch losses are
\begin{align}
\mathcal{L}_{\mathrm{seq}}
&=
\frac{1}{|\mathcal{G}|}
\sum_{i\in\mathcal{G}}
D_{\mathrm{KL}}
\left(
q(s_i^{t-1}\mid s_i^t,s_i^0)
\,\middle\|\,
p_\theta^{\mathrm{seq}}
(s_i^{t-1}\mid R_t,C,t)
\right),
\label{eq:seq-denoise}
\\
\mathcal{L}_{\mathrm{pos}}
&=
\frac{1}{|\mathcal{G}|}
\sum_{i\in\mathcal{G}}
\left\|
\epsilon_i^{\mathrm{pos}}
-
\hat{\epsilon}_{\theta,i}^{\mathrm{pos}}(R_t,C,t)
\right\|_2^2,
\label{eq:pos-denoise}
\\
\mathcal{L}_{\mathrm{ori}}
&=
\frac{1}{|\mathcal{G}|}
\sum_{i\in\mathcal{G}}
\left\|
(O_i^0)^\top
\hat{O}_{\theta,i}^{t-1}(R_t,C,t)
-
I_3
\right\|_F^2.
\label{eq:ori-denoise}
\end{align}
Here, $q(s_i^{t-1}\mid s_i^t,s_i^0)$ is the categorical diffusion posterior,
$\epsilon_i^{\mathrm{pos}}$ is the Gaussian noise applied to the
C$\alpha$ coordinate, and $\hat{O}_{\theta,i}^{t-1}$ is the predicted
denoised orientation. We use unit weights,
$\lambda_{\mathrm{pos}}=\lambda_{\mathrm{ori}}=1$, and retain the complete
objective as the offline denoising anchor during ABOPD post-training.

The reference-derived state distribution underlying Eq.~\eqref{eq:denoise} differs from the distribution encountered during generation. At inference, a reverse trajectory
\[
R_K^\theta\rightarrow R_{K-1}^\theta\rightarrow\cdots\rightarrow R_0^\theta
\]
recursively conditions each transition on previous model outputs. Consequently, even a model trained accurately on $q_t(R_t\mid R_0,\mathcal{G})$ may receive states outside this reference-derived distribution after its own errors accumulate. ABOPD addresses this mismatch by adding teacher supervision directly on states visited along the student's reverse trajectory.

\subsection{Hybrid Pretraining and Backbone-Aware Teacher Adaptation}

\paragraph{Hybrid pretraining.}

The released DiffAb protocol trains separate models for single-CDR and multi-CDR co-design \citep{luo2022diffab}. We instead pretrain a single H-DiffAb model across single-CDR, multi-CDR, and all-six-CDR target masks. At each update, a target set $\mathcal{G}\sim p_{\mathrm{hyb}}(\mathcal{G})$ is sampled; only residues in $\mathcal{G}$ are noised and reconstructed, while the remaining antibody and antigen residues provide structural context. The objective is
\begin{equation}
\label{eq:hyb-pretrain}
\mathcal{L}_{\mathrm{hyb}}
=
\mathbb{E}_{\mathcal{G},t,R_t\sim q_t(\cdot\mid R_0,\mathcal{G})}
\left[
\mathcal{L}_{\mathrm{denoise}}
\bigl(f_\theta(R_t,C,t),R_0;\mathcal{G}\bigr)
\right].
\end{equation}
We retain the same target-mask distribution during teacher adaptation and student post-training. The H-DiffAb pretraining budget is approximately $8\times$ that of a default DiffAb checkpoint; this larger budget should be considered when interpreting its improvement over DiffAb. The resulting parameters $\theta_0$ initialize the teacher, the ABOPD student, and all post-training controls, so comparisons among the post-training methods share the same starting point. The sampling probabilities and training configuration are reported in Appendix~\ref{app:training-details} and summarized in Table~\ref{tab:training-details}.

\paragraph{Backbone-aware teacher adaptation.}
The teacher augments H-DiffAb with a privileged descriptor $B$ constructed
from native target geometry. For each target residue, $B$ contains
C$\alpha$-centered Cartesian coordinates and validity masks for the N,
C$\alpha$, C, O, and observed C$\beta$ atoms, together with
target-to-complex C$\alpha$ distance features. Two lightweight residual MLP
adapters encode the residue- and pair-level descriptors and add them to the
corresponding H-DiffAb encoder embeddings. The enriched embeddings and the
unchanged noisy state $R_t$ are then processed by the original EpsilonNet, so
the privileged geometry conditions the denoising prediction without replacing
the diffusion state. Explicit target residue labels and atoms beyond
C$\beta$ are excluded, although the observed C$\beta$ geometry and validity
mask can carry limited sequence-correlated information. Exact descriptor
encodings and adapter architectures are provided in
Appendix~\ref{app:teacher-inputs}.

We initialize the H-DiffAb backbone from $\theta_0$, randomly initialize both
adapters, and jointly optimize the full teacher on reference-derived states:
\begin{equation}
\label{eq:teacher-adapt}
\mathcal{L}_{\mathrm{teacher}}
=
\mathbb{E}_{\mathcal{G},t,R_t\sim q_t(\cdot\mid R_0,\mathcal{G})}
\left[
\mathcal{L}_{\mathrm{denoise}}
\bigl(f_\phi(R_t,C,B,t),R_0;\mathcal{G}\bigr)
\right].
\end{equation}
After adaptation, the resulting parameters $\phi^\star$ are frozen, and $B$
is used only when querying the teacher during student post-training.

\subsection{ABOPD: On-Policy Distillation}
\label{sec:abopd}

\paragraph{Student-visited rollout states.}
The student $f_\theta$ is initialized with $\theta_0$ and receives only $(R_t,C,t)$. For each training batch, the current student generates a reverse-denoising trajectory using the inference-time sampler:
\[
R_K^S\rightarrow R_{K-1}^S\rightarrow\cdots\rightarrow R_0^S,
\qquad
R_{0:K}^S\sim p_\theta^{\mathrm{rollout}}(\cdot\mid C,\mathcal{G}).
\]

Each rollout starts at schedule index $K=80$, with the generated residues
sampled from the factorized initialization distribution used by DiffAb
\citep{luo2022diffab}: amino-acid types are sampled uniformly over 20
categories, normalized C$\alpha$ coordinates are sampled from $\mathcal{N}(0,I)$, and
residue orientations are sampled uniformly over $\mathrm{SO}(3)$. The rollout follows
$80\rightarrow79\rightarrow\cdots\rightarrow0$ on the original 100-step
schedule, while the antibody--antigen context remains fixed.
The rollout is performed without gradients, and the cached states are
detached before distillation. We cache the states at
$\mathcal{T}=\{80,70,60,50,40,30,20,10,5\}$; this training timestep set is
distinct from the eight-timestep grid used only for the mechanism analysis in
Section~\ref{sec:why-opd}.
Thus, ``on-policy'' refers only to the fact that distillation states are drawn from the current student's own inference process. It does not imply policy-gradient optimization.

At every cached state, the student and frozen teacher are evaluated on exactly the same $R_t^S$. For compactness, denote their coordinate predictions by
\begin{equation}
\label{eq:student-teacher-query}
\hat\epsilon_{S,t}^{\mathrm{pos}}
:=\hat\epsilon_\theta^{\mathrm{pos}}(R_t^S,C,t),
\qquad
\hat\epsilon_{T,t}^{\mathrm{pos}}
:=\hat\epsilon_{\phi^\star}^{\mathrm{pos}}(R_t^S,C,B,t).
\end{equation}
The teacher remains frozen throughout student optimization, and its predictions are treated as fixed structural targets; gradients propagate only through the student branch. Although $f_{\phi^\star}$ is fixed, its target is state-dependent because it is recomputed for each student-visited $R_t^S$.

\paragraph{From reverse KL to coordinate transition matching.}
Let $P_{S,t}$ and $P_{T,t}$ denote the student and teacher one-step reverse kernels evaluated at the same student-visited state. At each selected rollout timestep $t\in\mathcal{T}$, the local OPD target follows the reverse-KL direction used for student-generated diffusion trajectories \citep{li2026diffusionopd}:
\begin{equation}
\label{eq:generic-opd}
\ell_t^{\mathrm{OPD}}
=
D_{\mathrm{KL}}\!\left(
P_{S,t}(\cdot\mid R_t^S,C,\mathcal{G})
\,\middle\|\,
P_{T,t}(\cdot\mid R_t^S,C,B,\mathcal{G})
\right).
\end{equation}
The student kernel appears first because distillation is evaluated on states
visited by the student, following the reverse-KL OPD convention.

We instantiate Eq.~\eqref{eq:generic-opd} using the DDPM coordinate transition adopted by DiffAb \citep{ho2020ddpm,luo2022diffab}. This Gaussian transition is inherited from the coordinate sampler rather than introduced by ABOPD. Conditioned on a shared student-visited state, model $j\in\{S,T\}$ induces the one-step reverse kernel
\begin{equation}
\label{eq:coordinate-reverse-kernel}
P_{j,t}^{\mathrm{pos}}
=
\mathcal{N}\!\left(\mu_{j,t}^{\mathrm{pos}},\sigma_t^2I\right),
\qquad
\mu_{j,t}^{\mathrm{pos}}
=
a_t x_t^S+b_t\hat\epsilon_{j,t}^{\mathrm{pos}}.
\end{equation}
Given $x_t^S$, the model prediction is deterministic and the only transition
randomness is the sampler's Gaussian noise. The coefficients $a_t$, $b_t$, and
$\sigma_t^2$ are determined by the shared DiffAb schedule. Student and Teacher
therefore have the same covariance and differ only through their
coordinate-noise predictions. In particular, their mean difference is
\[
\begin{aligned}
\mu_{S,t}^{\mathrm{pos}}-\mu_{T,t}^{\mathrm{pos}}
&=
\left(a_t x_t^S+b_t\hat\epsilon_{S,t}^{\mathrm{pos}}\right)
-
\left(a_t x_t^S+b_t\hat\epsilon_{T,t}^{\mathrm{pos}}\right) \\
&=
b_t\left(
\hat\epsilon_{S,t}^{\mathrm{pos}}
-\hat\epsilon_{T,t}^{\mathrm{pos}}
\right).
\end{aligned}
\]
Substituting this difference into the KL between the shared-covariance Gaussian kernels in Eq.~\eqref{eq:coordinate-reverse-kernel} reduces the coordinate component of Eq.~\eqref{eq:generic-opd} to
\begin{equation}
\label{eq:coordinate-kl}
D_{\mathrm{KL}}\!\left(
P_{S,t}^{\mathrm{pos}}
\middle\|
P_{T,t}^{\mathrm{pos}}
\right)
=
\frac{b_t^2}{2\sigma_t^2}
\left\|
\hat\epsilon_{S,t}^{\mathrm{pos}}
-\hat\epsilon_{T,t}^{\mathrm{pos}}
\right\|_2^2.
\end{equation}
To remove the explicit schedule-dependent KL prefactor and simplify implementation, we drop the factor
$b_t^2/(2\sigma_t^2)$ and assign equal weight to all selected timesteps, Let $\mathcal D$ denote the training-data distribution over $(C,R_0,B)$.:
\begin{equation}
\label{eq:opd-pos}
\mathcal{L}_{\mathrm{pos}}^{\mathrm{OPD}}
=
\mathbb{E}_{\substack{
(C,R_0,B)\sim\mathcal{D},\;\mathcal{G}\sim p_{\mathrm{hyb}},\\
R_{0:K}^S\sim p_{\theta}^{\mathrm{rollout}}(\cdot\mid C,\mathcal{G})
}}
\left[
\frac{1}{|\mathcal{T}|\,|\mathcal{G}|}
\sum_{t\in\mathcal{T}}
\sum_{i\in\mathcal{G}}
\left\|
\hat\epsilon_{S,t,i}^{\mathrm{pos}}
-\hat\epsilon_{T,t,i}^{\mathrm{pos}}
\right\|_2^2
\right].
\end{equation}
Equation~\eqref{eq:opd-pos} is a uniformly reweighted
coordinate-transition regression objective. It retains the per-state
teacher-matching target induced by Eq.~\eqref{eq:coordinate-kl}, while
omitting the schedule-dependent KL weights. Because rollout states are
detached, gradients pass only through the student predictions at the sampled
states, not through the rollout distribution. Thus, reverse KL motivates the
state distribution and distillation direction, whereas
Eq.~\eqref{eq:opd-pos} defines the practical ABOPD update.

We distill only the coordinate branch because it provides the most direct on-policy correction to CDR geometry. The C$\alpha$ trace controls framework anchoring and loop placement at the antigen interface; through Eq.~\eqref{eq:coordinate-reverse-kernel}, coordinate errors directly alter the next loop state and thereby the geometric context of subsequent denoising steps. Sequence predictions do not themselves specify a spatial correction, whereas residue orientations describe local frames on the current coordinate scaffold and cannot alone correct translational drift in loop placement. The Backbone-aware Teacher improves coordinate denoising and backbone recovery
without improving sequence recovery
(Table~\ref{tab:teacher-validation}). Coordinate matching also provides the
best trade-off in the distillation-target ablation
(Fig.~\ref{fig:transition-component-ablation}). We therefore apply OPD only to C$\alpha$ coordinates, while sequence and orientation remain supervised by the denoising anchor below.

\paragraph{Offline denoising anchor.}
On-policy coordinate supervision alone does not preserve the joint sequence,
coordinate, and orientation objective learned during pretraining. We therefore retain the original denoising loss on an independently sampled reference-derived state:
\begin{equation}
\label{eq:anchor}
\mathcal{L}_{\mathrm{anchor}}
=
\mathbb{E}_{\substack{
(C,R_0)\sim\mathcal{D},\;\mathcal{G}\sim p_{\mathrm{hyb}},\\
t\sim\mathcal{U}\{1,\ldots,K\},\;R_t\sim q_t(\cdot\mid R_0,\mathcal{G})
}}
\left[
\mathcal{L}_{\mathrm{denoise}}
\bigl(f_\theta(R_t,C,t),R_0;\mathcal{G}\bigr)
\right].
\end{equation}
The final objective is
\begin{equation}
\label{eq:abopd-objective}
\boxed{
\mathcal{L}_{\mathrm{ABOPD}}
=
\mathcal{L}_{\mathrm{anchor}}
+\beta\mathcal{L}_{\mathrm{pos}}^{\mathrm{OPD}}
}.
\end{equation}
Here, $\beta$ controls the relative strength of on-policy coordinate
supervision, and we set $\beta=0.6$. ABOPD is therefore a mixed-state-distribution objective:
$\mathcal{L}_{\mathrm{anchor}}$ is evaluated on $q_t(R_t\mid R_0,\mathcal{G})$,
whereas $\mathcal{L}_{\mathrm{pos}}^{\mathrm{OPD}}$ is evaluated on
$p_\theta^{\mathrm{rollout}}(\cdot\mid C,\mathcal{G})$. We maintain an exponential moving average (EMA) copy of the student for
validation and generation, while the current non-EMA student produces all
training rollouts and the teacher remains frozen. Unless otherwise stated, reported ABOPD results use the EMA student $f_{\bar\theta}$, whereas SFT and offline-distillation controls use their directly optimized parameters. The EMA analysis is reported in Appendix~\ref{app:ema-analysis}. The complete post-training procedure is summarized in Algorithm~\ref{alg:abopd-training}.

\refstepcounter{methodalgorithm}
\label{alg:abopd-training}
\begin{tcolorbox}[
title={Algorithm \themethodalgorithm: ABOPD with On-Policy Coordinate Transition Matching},
colback=white,
colframe=black!25,
boxrule=0.5pt,
arc=1mm,
left=4pt,
right=4pt,
top=4pt,
bottom=4pt,
float=t,
]
\footnotesize
\begin{algorithmic}[1]
\Require $f_{\theta_0}$; frozen teacher $f_{\phi^\star}$; data $\mathcal{D}$;
$\mathcal{T}$; $\beta$; EMA decay $\rho$

\State $\theta\gets\theta_0,\quad \bar{\theta}\gets\theta$

\For{each optimization step}
    \State $(C,R_0,B)\sim\mathcal{D}$,
    $\mathcal{G}\sim p_{\mathrm{hyb}}(\mathcal{G})$

    \State $t_a\sim\mathcal{U}\{1,\ldots,K\}$,
    $R_{t_a}^{\mathrm{ref}}\sim q_{t_a}(\cdot\mid R_0,\mathcal{G})$
    \State $\mathcal{L}_{\mathrm{anchor}}
    \gets
    \mathcal{L}_{\mathrm{denoise}}
    \bigl(f_\theta(R_{t_a}^{\mathrm{ref}},C,t_a),R_0;\mathcal{G}\bigr)$

    \State With gradients disabled, $\{R_\tau^S\}_{\tau\in\mathcal{T}}
    \gets \operatorname{Rollout}(f_\theta,C,\mathcal{G})$

    \ForAll{$\tau\in\mathcal{T}$}
        \State $\hat{\epsilon}_{S,\tau}^{\mathrm{pos}}
        \gets f_\theta^{\mathrm{pos}}(R_\tau^S,C,\tau)$
        \State $\hat{\epsilon}_{T,\tau}^{\mathrm{pos}}
        \gets f_{\phi^\star}^{\mathrm{pos}}(R_\tau^S,C,B,\tau)$
    \EndFor

    \State Compute $\mathcal{L}_{\mathrm{pos}}^{\mathrm{OPD}}$
    using Eq.~\eqref{eq:opd-pos} and set
    $\mathcal{L}\gets
    \mathcal{L}_{\mathrm{anchor}}
    +\beta\mathcal{L}_{\mathrm{pos}}^{\mathrm{OPD}}$

    \State $\theta\gets
    \operatorname{Update}(\theta,\nabla_\theta\mathcal{L});
    \quad
    \bar{\theta}\gets
    \operatorname{EMA}_{\rho}(\bar{\theta},\theta)$
\EndFor

\State \Return $f_{\bar{\theta}}$
\end{algorithmic}
\end{tcolorbox}

\paragraph{Post-training controls.}
SFT continues the reference-state denoising objective without using the teacher or student rollout states:
\begin{equation}
\label{eq:sft-relation}
\mathcal{L}_{\mathrm{SFT}}
=
\mathbb{E}_{\substack{
(C,R_0)\sim\mathcal{D},\;\mathcal{G}\sim p_{\mathrm{hyb}},\\
t\sim\mathcal{U}\{1,\ldots,K\},\;R_t\sim q_t(\cdot\mid R_0,\mathcal{G})
}}
\left[
\mathcal{L}_{\mathrm{denoise}}
\bigl(f_\theta(R_t,C,t),R_0;\mathcal{G}\bigr)
\right].
\end{equation}
Thus, SFT is identical in form to $\mathcal{L}_{\mathrm{anchor}}$ but is optimized without an OPD term. Offline distillation instead applies the same frozen teacher's coordinate target to reference-derived states. Below, $\hat\epsilon_S^{\mathrm{pos}}(R_t)$ and $\hat\epsilon_T^{\mathrm{pos}}(R_t)$ abbreviate the student and teacher predictions under conditioning $C$ and $(C,B)$, respectively:
\begin{equation}
\label{eq:offline-distillation}
\mathcal{L}_{\mathrm{pos}}^{\mathrm{offline}}
=
\mathbb{E}_{\substack{
(C,R_0,B)\sim\mathcal{D},\;\mathcal{G}\sim p_{\mathrm{hyb}},\\
\{R_t^{\mathrm{ref}}\sim q_t(\cdot\mid R_0,\mathcal{G})\}_{t\in\mathcal{T}}
}}
\left[
\frac{1}{|\mathcal{T}|\,|\mathcal{G}|}
\sum_{t\in\mathcal{T}}
\sum_{i\in\mathcal{G}}
\left\|
\hat\epsilon_{S,t,i}^{\mathrm{pos}}(R_t^{\mathrm{ref}})
-\hat\epsilon_{T,t,i}^{\mathrm{pos}}(R_t^{\mathrm{ref}})
\right\|_2^2
\right].
\end{equation}
Offline distillation uses the same Teacher target, selected timestep set, and uniform timestep weighting as ABOPD, but applies the target to reference-derived states. It does not include $\mathcal{L}_{\mathrm{anchor}}$. All post-training controls use the same H-DiffAb initialization, training data, and number of optimizer updates; they differ in their supervision and state distributions.

\section{Experiments}
The experiments address three questions: whether the Backbone-aware Teacher
provides more accurate geometric targets, whether ABOPD improves CDR generation
across two complementary benchmarks, and whether on-policy supervision provides
gains beyond SFT and offline distillation. All post-training controls start
from the same H-DiffAb initialization.

\subsection{Setup}

\paragraph{Data.}
Following \citet{luo2022diffab}, we construct the data split from the
SAbDab database \citep{dunbar2014sabdab}. We exclude structures with resolution
worse than 4~\AA{} and antibodies targeting non-protein antigens, then cluster
the remaining antibody--antigen complexes by CDR-H3 sequence at 50\% identity.
Five clusters are held out for testing, and the
remaining clusters are used for training.

\paragraph{Benchmarks.}
We evaluate two complementary generation settings. Simultaneous multi-CDR
redesign jointly generates the sequences and structures of all six CDRs for
19 held-out SAbDab antibody--antigen complexes. RAbD CDR-H3 generation evaluates
the most diverse and conformationally flexible heavy-chain CDR loop across
60 target cases \citep{adolf2018rabd}.

\paragraph{Metrics.}
\textbf{AAR} (\%) measures amino-acid recovery over generated CDR residues; we
report it as a conventional metric but interpret it cautiously because it does
not necessarily reflect overall design quality \citep{kong2024PepGlad}.
Structural fidelity is evaluated using C$\alpha$ \textbf{RMSD} in the fixed
complex coordinate frame without Kabsch alignment, \textbf{lDDT}, and
heavy-chain \textbf{TM-Score}. Following \citet{wu2026proteor1}, we compute
antibody-global lDDT on PyRosetta-packed all-atom structures.
Interface quality is assessed using \textbf{ddG} and \textbf{DockQ}. We define
ddG as $dG_{\mathrm{gen}}-dG_{\mathrm{ref}}$, where lower values indicate a
more favorable interface-energy change
\citep{leaverfay2011rosetta3,alford2017rosetta}. In simultaneous multi-CDR
redesign, ddG is computed over the complete antibody--antigen interface.
\footnote{Before aggregating ddG, we exclude catastrophic Rosetta relaxation
or energy-evaluation failures, affecting 0.433\% of simultaneous multi-CDR
samples.}
DockQ measures agreement with the native protein--protein interface
\citep{basu2016dockq} and is computed between the antigen and complete antibody
heavy chain following \citet{wu2026proteor1}.
Geometric realism is evaluated using \textbf{Clash$_{\mathrm{in}}$},
\textbf{Clash$_{\mathrm{out}}$}, and \textbf{JSD$_{\mathrm{bb}}$}. The clash
metrics measure C$\alpha$-level steric conflicts within the generated CDRs and
between generated CDRs and the antigen, respectively, using a 3.6574~\AA{}
cutoff and excluding covalently adjacent CDR residues
\citep{kong2025unimomo}. JSD$_{\mathrm{bb}}$ is the Jensen--Shannon divergence
between generated and reference backbone dihedral-angle distributions. All
three metrics are computed on raw generated structures before relaxation.

\paragraph{Sampling protocol.}
For each target and sampling seed, we generate 100 designs and aggregate results
for our evaluated variants over three fixed seeds. Baseline implementation and
evaluation details are provided in Appendix~\ref{app:baseline-methods}.

\subsection{Backbone-Aware Teacher Validation}

Because the Teacher supplies training-time distillation targets rather than
serving as the final generator, we evaluate H-DiffAb, Teacher w/o Backbone,
and the Backbone-aware Teacher on identical reference-derived noisy states
$(R_t,C,t)$. The two adapted variants follow the same procedure and differ
only in access to privileged backbone descriptors.
Table~\ref{tab:teacher-validation} shows that the Backbone-aware Teacher
reduces coordinate MSE from 0.2019 to 0.1503 and orientation loss from 0.1445
to 0.1324 relative to H-DiffAb, while its sequence accuracy remains lower,
localizing the benefit to geometric prediction. The no-backbone continuation
control improves sampled CDR RMSD despite exhibiting poorer one-step denoising accuracy,
showing that sampled-generation quality alone is insufficient for selecting a
distillation teacher. We therefore freeze the Backbone-aware Teacher, which
provides the most accurate geometric denoising targets, for subsequent student
distillation. Additional adaptation variants and timestep-resolved results are
reported in Appendix~\ref{app:teacher-analysis}.

\begin{table}[H]
\centering
\caption{\textbf{Backbone-aware Teacher validation.}
Coord.~MSE, Ori.~loss, and Seq.~Acc.\ evaluate one-step denoising on shared
reference-derived states, whereas AAR and RMSDs characterize samples generated
by the same variants. CDR BB RMSD is computed over the backbone atoms
N, C$\alpha$, and C in the fixed complex coordinate frame. Best and second-best
values are shown in \textbf{bold} and \protect\underline{underlined},
respectively.}
\label{tab:teacher-validation}
\scriptsize
\setlength{\tabcolsep}{3.5pt}
\begin{tabular}{lcccccc}
\toprule
Variant
& Coord. MSE $\downarrow$
& Ori. loss $\downarrow$
& Seq. Acc. $\uparrow$
& AAR (\%) $\uparrow$
& CDR C$\alpha$ RMSD (\AA) $\downarrow$
& CDR BB RMSD (\AA) $\downarrow$ \\
\midrule
H-DiffAb
& \underline{0.2019}
& \underline{0.1445}
& \textbf{0.7382}
& \textbf{60.56}
& 1.4863
& 1.5298 \\
Teacher w/o Backbone
& 0.2212
& 0.1662
& 0.7105
& 57.92
& \underline{1.0520}
& \underline{1.1428} \\
Backbone-aware Teacher
& \textbf{0.1503}
& \textbf{0.1324}
& \underline{0.7246}
& \underline{59.26}
& \textbf{0.4855}
& \textbf{0.6642} \\
\bottomrule
\end{tabular}
\end{table}

\subsection{Main Results}
\paragraph{Simultaneous multi-CDR redesign.}
Table~\ref{tab:all-cdr-joint} shows that, relative to H-DiffAb, ABOPD
reduces RMSD across all six CDRs, with the largest improvement on CDR-H3,
from 2.94~\AA{} to 2.48~\AA{}. It also reduces intra-CDR and CDR--antigen
clash rates from 0.93\% and 1.58\% to 0.42\% and 0.67\%, respectively,
while maintaining a similar JSD$_{\mathrm{bb}}$. Together, these results
indicate more accurate multi-loop geometry while preserving
backbone-dihedral realism. Single-CDR redesign results are reported in
Appendix~\ref{app:per-cdr-redesign}.

\begin{table}[H]
\centering
\caption{\textbf{Simultaneous multi-CDR redesign.}
H3 is highlighted as the most flexible and challenging CDR.
For entries containing $\pm$, we report the mean and 95\% confidence interval over three sampling seeds. Best and second-best results are shown in \textbf{bold} and
\protect\underline{underlined}, respectively.}
\label{tab:all-cdr-joint}
\scriptsize
\setlength{\tabcolsep}{2pt}
\begin{adjustbox}{max width=\linewidth}
\begin{tabular}{lcccccccccccc}
\toprule
Method & AAR (\%) $\uparrow$ & lDDT $\uparrow$ & ddG $\downarrow$ & \multicolumn{6}{c}{Per-CDR RMSD (\AA) $\downarrow$} & \multicolumn{3}{c}{Geometric Realism $\downarrow$} \\
\cmidrule(lr){5-10}\cmidrule(lr){11-13}
 & & & & H1 & H2 & \cellcolor{h3colblue}H3 & L1 & L2 & L3 & Clash$_{\mathrm{in}}$ (\%) & Clash$_{\mathrm{out}}$ (\%) & JSD$_{\mathrm{bb}}$ \\
\midrule
dyMEAN~\citep{kong2023dymean} & 60.07 & 0.8029 & 60.03 & 1.65 & 1.47 & \cellcolor{h3colblue}6.15 & 1.58 & 1.23 & 1.59 & -- & -- & -- \\
AbX~\citep{zhu2024Abx} & 26.52 & 0.8365 & 22.65 & 1.46 & 1.49 & \cellcolor{h3colblue}6.01 & 3.19 & 2.78 & 1.30 & -- & -- & -- \\
PPIFlow~\citep{ppiflow2026} & -- & 0.3075 & -- & 2.17 & 1.68 & \cellcolor{h3colblue}5.41 & 3.64 & 2.73 & 4.89 & -- & -- & -- \\
IgGM~\citep{iggm2025} & 28.49 & 0.4234 & 23.93 & 1.73 & 1.55 & \cellcolor{h3colblue}4.37 & 1.62 & 1.51 & 1.71 & 25.63 & \underline{1.45} & 0.2900 \\
\midrule
DiffAb & 52.06{\scriptsize$\pm$0.29} & 0.8525{\scriptsize$\pm$0.0006} & 27.727{\scriptsize$\pm$25.835} & 0.79{\scriptsize$\pm$0.05} & 0.77{\scriptsize$\pm$0.26} & \cellcolor{h3colblue}3.41{\scriptsize$\pm$0.18} & 1.15{\scriptsize$\pm$0.16} & 1.11{\scriptsize$\pm$0.32} & 1.12{\scriptsize$\pm$0.24} & 1.40 & 1.87 & 0.2345 \\
H-DiffAb & \underline{60.27{\scriptsize$\pm$0.11}} & \underline{0.8670{\scriptsize$\pm$0.0004}} & \underline{22.431{\scriptsize$\pm$12.142}} & \underline{0.68{\scriptsize$\pm$0.02}} & \underline{0.66{\scriptsize$\pm$0.18}} & \cellcolor{h3colblue}\underline{2.94{\scriptsize$\pm$0.22}} & \underline{0.95{\scriptsize$\pm$0.20}} & \underline{1.04{\scriptsize$\pm$0.23}} & \underline{1.01{\scriptsize$\pm$0.23}} & \underline{0.93} & 1.58 & \textbf{0.2308} \\
ABOPD & \textbf{60.73{\scriptsize$\pm$0.10}} & \textbf{0.8710{\scriptsize$\pm$0.0004}} & \textbf{20.574{\scriptsize$\pm$10.627}} & \textbf{0.63{\scriptsize$\pm$0.02}} & \textbf{0.64{\scriptsize$\pm$0.20}} & \cellcolor{h3colblue}\textbf{2.48{\scriptsize$\pm$0.16}} & \textbf{0.90{\scriptsize$\pm$0.21}} & \textbf{1.00{\scriptsize$\pm$0.17}} & \textbf{0.97{\scriptsize$\pm$0.20}} & \textbf{0.42} & \underline{0.67} & \underline{0.2325} \\
\bottomrule
\end{tabular}
\end{adjustbox}
\end{table}

\paragraph{RAbD CDR-H3 evaluation.}
On the external RAbD benchmark, ABOPD reduces CDR-H3 RMSD from
2.37~\AA{} to 1.95~\AA{}, while AAR increases only modestly from
36.35\% to 37.31\% (Table~\ref{tab:rabd-hcdr3}), identifying improved
backbone recovery as its primary benefit. The accompanying gains in lDDT,
TM-score, and DockQ, together with lower intra-CDR and CDR--antigen clash
rates, indicate consistent improvements in local geometry, global heavy-chain
structure, and interface placement. Additional interface and side-chain
analyses are reported in Appendix~\ref{app:hcdr3-diagnostics}.

\begin{table}[t]
\centering
\caption{\textbf{RAbD CDR-H3 generation.}
For entries containing $\pm$, we report the mean and 95\% confidence interval over three sampling seeds. Methods marked with
\textsuperscript{*} use the pipeline IgFold $\Rightarrow$ HDOCK
$\Rightarrow$ CDR generation $\Rightarrow$ Rosetta side-chain packing
\citep{ruffolo2023igfold,yan2020hdock,leaverfay2011rosetta3}. Best and second-best results are shown in \textbf{bold} and
\protect\underline{underlined}, respectively.}
\label{tab:rabd-hcdr3}
\scriptsize
\setlength{\tabcolsep}{2pt}
\begin{tabular}{lcccccccc}
\toprule
Method & AAR (\%) $\uparrow$ & lDDT $\uparrow$ & TM-score $\uparrow$ & RMSD (\AA) $\downarrow$ & DockQ $\uparrow$ & Clash$_{\mathrm{in}}$ (\%) $\downarrow$ & Clash$_{\mathrm{out}}$ (\%) $\downarrow$ & JSD$_{\mathrm{bb}}$ $\downarrow$ \\
\midrule
RosettaAb\textsuperscript{*}~\citep{adolf2018rabd} & 32.31 & 0.8272 & 0.9717 & 17.70 & 0.14 & -- & -- & -- \\
MEAN\textsuperscript{*}~\citep{kong2022mean} & 37.38 & 0.8252 & 0.9688 & 17.30 & 0.16 & -- & -- & 0.5290 \\
GeoAB\textsuperscript{*}~\citep{lin2024geoab} & 40.02 & 0.8367 & 0.9695 & 15.43 & 0.19 & -- & -- & -- \\
dyMEAN~\citep{kong2023dymean} & \underline{41.84} & 0.8392 & 0.9718 & 8.10 & 0.41 & -- & -- & 0.5420 \\
DGENet~\citep{huang2026dgenet} & \textbf{42.67} & 0.8551 & 0.9747 & 7.19 & 0.43 & -- & -- & -- \\
BoltzGen~\citep{stark2025boltzgen} & 39.07 & 0.8372 & 0.9675 & 2.69 & 0.47 & -- & -- & -- \\
\midrule
DiffAb & 32.30{\scriptsize$\pm$0.63} & 0.9105{\scriptsize$\pm$0.0002} & 0.9772 & 2.79{\scriptsize$\pm$0.02} & 0.79 & \underline{3.18} & \underline{3.42} & 0.2622 \\
H-DiffAb & 36.35{\scriptsize$\pm$0.59} & \underline{0.9158{\scriptsize$\pm$0.0001}} & \underline{0.9809} & \underline{2.37{\scriptsize$\pm$0.02}} & \underline{0.81} & 3.46 & 3.69 & \textbf{0.2382} \\
ABOPD & 37.31{\scriptsize$\pm$0.21} & \textbf{0.9193{\scriptsize$\pm$0.0002}} & \textbf{0.9845} & \textbf{1.95{\scriptsize$\pm$0.02}} & \textbf{0.83} & \textbf{0.94} & \textbf{2.24} & \underline{0.2418} \\
\bottomrule
\end{tabular}
\end{table}

\begin{figure}[t]
\centering
\includegraphics[width=\linewidth]{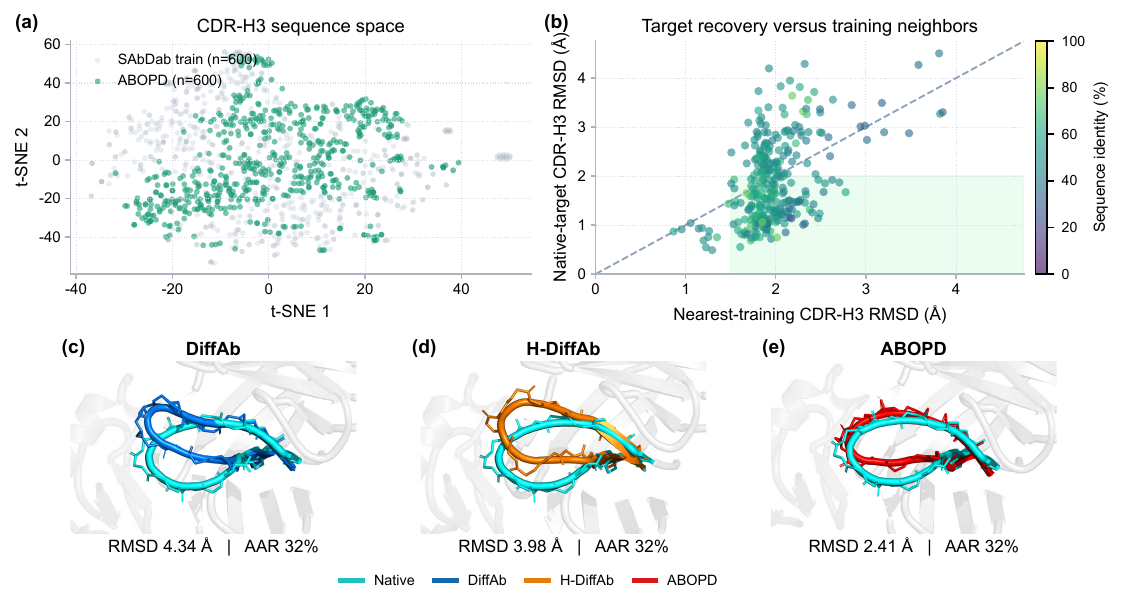}
\caption{
\textbf{(a)} t-SNE comparison of ABOPD and SAbDab training sequences.
\textbf{(b)} Native-target RMSD versus nearest-training CDR-H3 RMSD.
\textbf{(c--e)} Qualitative structural comparison with the native loop.}
\label{fig:generated-design-analysis}
\end{figure}

\paragraph{Sequence and structure analysis of generated CDR-H3 loops.}
Figures~\ref{fig:generated-design-analysis}(a) and~\ref{fig:generated-design-analysis}(b) show that ABOPD sequences
broadly overlap with SAbDab training CDRs in the t-SNE embedding, while the
generated structures are closer on average to their native targets than to
their nearest training neighbors. Together, these observations suggest that
the structural gains are not explained by simple nearest-neighbor retrieval.
In a representative case where all three methods attain the same AAR, ABOPD
better preserves framework anchoring and antigen-facing loop placement than
DiffAb and H-DiffAb
(Fig.~\ref{fig:generated-design-analysis}(c)--(e)).

\subsection{On-Policy Distillation: Controls and Mechanism}
\label{sec:why-opd}

\begin{figure}[t]
\centering
\includegraphics[width=0.9\linewidth]{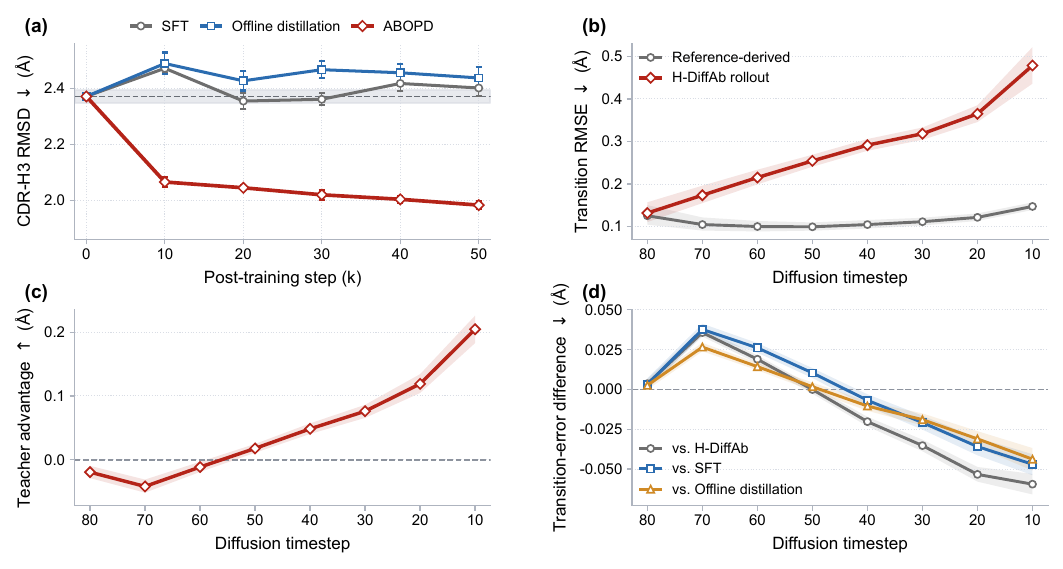}
\caption{
\textbf{(a)} CDR-H3 RMSD over post-training steps; the dashed line and
gray band denote the H-DiffAb initialization. The ABOPD curve uses EMA
checkpoints, whereas the control curves use directly optimized parameters.
\textbf{(b)} H-DiffAb C$\alpha$ transition error on reference-derived and
rollout states.
\textbf{(c)} Teacher advantage on rollout states, computed as paired
H-DiffAb error minus Teacher error; positive values favor the Teacher.
\textbf{(d)} Paired ABOPD error minus control error on the shared H-DiffAb
rollout-state bank; negative values favor ABOPD. Curves in
\textbf{(b--d)} are complex-level means after seed averaging; error bars
in \textbf{(a)} and shaded bands in \textbf{(b--d)} denote 95\%
confidence intervals computed across the three sampling seeds and by
complex-level bootstrap, respectively. Timesteps decrease from left to
right, corresponding to early-to-late reverse denoising.}
\label{fig:on-policy-controls-mechanism}
\end{figure}

\paragraph{Post-training controls.}
Starting from the same H-DiffAb initialization, we compare ABOPD with SFT,
which continues supervised denoising, and offline distillation, which applies
the same frozen Teacher and coordinate-transition target to reference-derived
states. Fig.~\ref{fig:on-policy-controls-mechanism}(a) shows that both controls
remain close to H-DiffAb, whereas ABOPD steadily reduces CDR-H3 RMSD. Thus,
neither continued denoising nor Teacher matching on reference-derived states
reproduces ABOPD's structural gain, consistent with the benefit of combining
the denoising anchor with Teacher supervision on rollout states. Although the
main ABOPD curve uses EMA checkpoints, the non-EMA student retains most of the
gain (Appendix~\ref{app:ema-analysis}, Table~\ref{tab:ema-ablation}), indicating
that EMA alone does not explain the improvement.

\paragraph{Timestep-level mechanism analysis.}
Using all 60 RAbD CDR-H3 complexes and three sampling seeds, we compare
reference-derived states from forward noising with states cached from H-DiffAb
rollouts at $t\in\{80,70,60,50,40,30,20,10\}$. At each timestep, H-DiffAb and
the frozen Teacher are evaluated on the same state, with the Teacher
additionally receiving the privileged descriptor $B$. Transition error is the
one-step C$\alpha$ RMSE to a native-directed transition-mean target, which
serves as a local reference rather than the exact reverse posterior on rollout
states. Teacher advantage is computed as H-DiffAb error minus Teacher error,
whereas the ABOPD--control difference is ABOPD error minus the corresponding
control error; positive and negative values favor the Teacher and ABOPD,
respectively.

H-DiffAb has higher transition error on rollout states than on
reference-derived states throughout denoising, with the gap widening toward
later timesteps (Fig.~\ref{fig:on-policy-controls-mechanism}(b)). Teacher
advantage becomes positive at $t=50$, whereas ABOPD attains lower transition
error than all controls at $t=40$--$10$
(Fig.~\ref{fig:on-policy-controls-mechanism}(c)--(d)). Together, these temporal
patterns support a stage-dependent mechanism in which rollout-state
supervision becomes most useful during later geometric refinement
(Fig.~\ref{fig:on-policy-controls-mechanism}(b)--(d)).

\subsection{Distillation Target Ablation}
\label{sec:target-ablation}

\begin{figure}[H]
\centering
\includegraphics[width=0.8\linewidth]{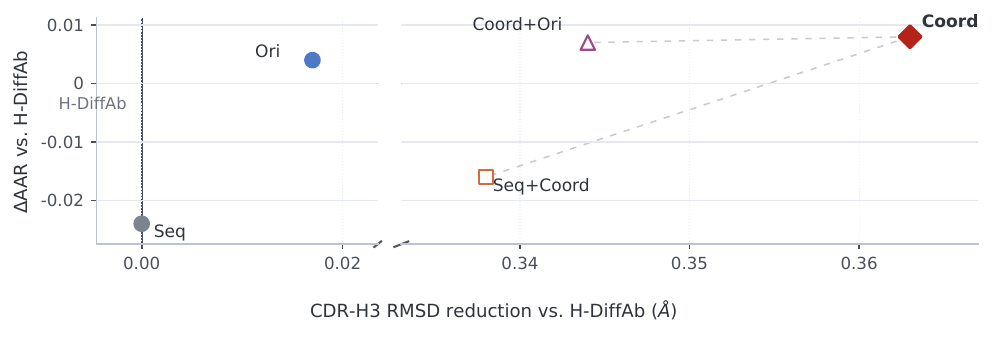}
\caption{Distillation target ablation on RAbD CDR-H3. Each point reports
the AAR gain and CDR-H3 RMSD reduction relative to H-DiffAb; the x-axis is
broken for readability.}
\label{fig:transition-component-ablation}
\end{figure}

We compare sequence, coordinate, orientation, and combined distillation
targets under the same H-DiffAb initialization, Teacher, rollout budget, and
sampling protocol. Sequence-only matching decreases AAR without improving
RMSD, whereas orientation-only matching yields only a limited structural gain
(Fig.~\ref{fig:transition-component-ablation}). All variants containing the
coordinate target substantially reduce RMSD, but coordinate-only matching
achieves the largest RMSD reduction and AAR gain; adding sequence or
orientation provides no further improvement. Together with the Teacher's
advantage in geometric denoising but not sequence accuracy
(Table~\ref{tab:teacher-validation}), these results support the coordinate-only
distillation target adopted by ABOPD.

\section{Conclusion}
We introduce ABOPD, an on-policy distillation framework that improves the
structural accuracy of antibody CDR design by transferring privileged
geometric supervision to states visited along the student's own trajectories.
ABOPD reduces RAbD CDR-H3 RMSD from 2.37~\AA{} to 1.95~\AA{} and improves
RMSD across all six CDRs in simultaneous redesign, outperforming SFT and
offline distillation controls. Timestep-level analysis reveals a
stage-dependent correspondence: the gap between rollout and reference-derived
states widens during denoising, while Teacher corrections become most useful
during later geometric refinement. These findings highlight that effective
structural post-training depends not only on the supervision signal, but also
on the generated states to which it is applied. ABOPD therefore provides a
complementary direction to larger-scale pretraining and new generative
architectures for more reliable antibody and protein design.

\clearpage

\bibliographystyle{unsrtnat}
\bibliography{main}

\clearpage
\appendix
\renewcommand{\thesection}{\Alph{section}}
\renewcommand{\thesubsection}{\thesection.\arabic{subsection}}
\renewcommand{\thesubsubsection}{\thesubsection.\arabic{subsubsection}}

\section{Supplementary Experimental Results}
\label{app:supplementary-results}

\subsection{Additional Teacher Evaluation}
\label{app:teacher-analysis}

\begin{table}[H]
\centering
\caption{\textbf{Teacher training variants.}
The adapter-only Teacher achieves the highest AAR, whereas the backbone-aware Teacher minimizes both CDR geometry errors and provides the frozen structural target used by ABOPD.
Both RMSDs evaluate generated CDR geometry rather than the fixed framework.
Best and second-best results are shown in \textbf{bold} and
\protect\underline{underlined}, respectively.}
\label{tab:teacher-training-variants}
\small
\begin{tabular}{lccc}
\toprule
Variant & AAR (\%) $\uparrow$ & CDR C$\alpha$ RMSD (\AA) $\downarrow$ & CDR BB RMSD (\AA) $\downarrow$ \\
\midrule
H-DiffAb & \underline{60.56} & 1.4863 & 1.5298 \\
Teacher w/o Backbone & 57.92 & 1.0520 & 1.1428 \\
Adapter-only Teacher & \textbf{61.48} & \underline{0.9133} & \underline{0.9962} \\
Backbone-aware Teacher & 59.26 & \textbf{0.4855} & \textbf{0.6642} \\
\bottomrule
\end{tabular}
\end{table}

\begin{figure}[H]
\centering
\includegraphics[width=0.9\linewidth]{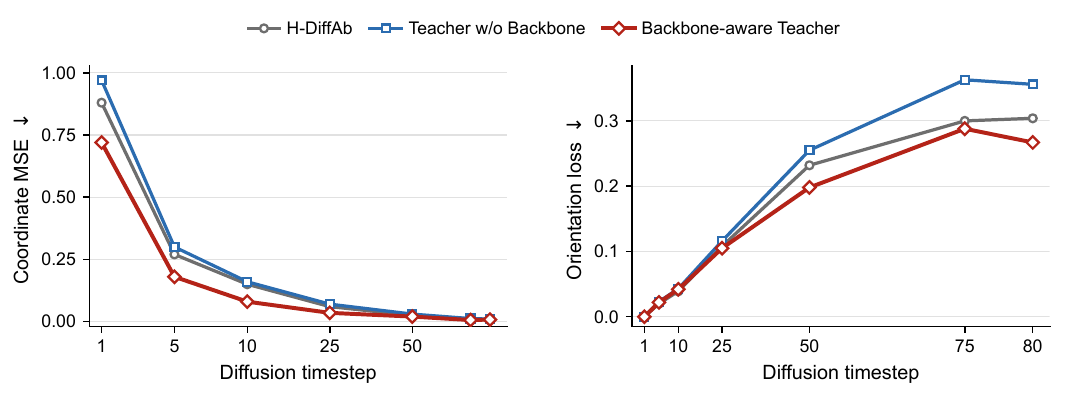}
\caption{\textbf{Timestep-resolved Teacher evaluation.}
On shared reference-derived noisy states, backbone-aware conditioning most clearly reduces coordinate MSE at low-index timesteps and orientation loss at higher-index timesteps.
Orientation loss is diagnostic only and is not distilled.}
\label{fig:backbone-aware-teacher-transition-probe}
\end{figure}

\subsection{Additional Post-Training Trajectories}
\label{app:training-trajectories}
\begin{figure}[H]
\centering
\includegraphics[width=0.8\linewidth]{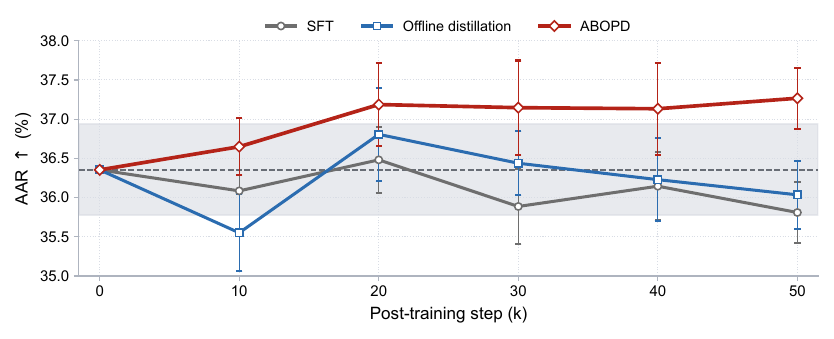}
\caption{\textbf{Post-training AAR trajectories on RAbD CDR-H3.}
ABOPD attains higher mean AAR than SFT, offline distillation, and the H-DiffAb initialization at every evaluated checkpoint after initialization.
Markers and error bars show the mean and 95\% confidence interval over three sampling seeds; the dashed line and gray band denote the H-DiffAb initialization.}
\label{fig:aar-training-trajectory}
\end{figure}

\subsection{Teacher-Correction Geometry Across Timesteps}
\label{app:timestep-mechanism}

The scalar Teacher advantage in
Fig.~\ref{fig:on-policy-controls-mechanism}(c) reveals whether the Teacher is
closer than H-DiffAb to the native-conditioned one-step target, but not how the
Teacher changes H-DiffAb's transition. We therefore treat the difference
between their predicted transition means as an additive correction. This view
is motivated by diffusion-guidance analyses that interpret differences between
denoiser predictions as corrections to a base vector field and characterize
such corrections through their directional relation to a reference denoising
signal and their norm across timesteps
\citep{karras2024autoguidance,rajabi2025tpg}. Accordingly, we ask whether the
Teacher correction points toward the native-conditioned transition and how
large that correction is. This analogy is diagnostic only: the correction is
not applied as inference-time guidance.

For a current coordinate $x_{t,i}$ and its corresponding native coordinate
$x_{0,i}$, we construct the native-directed transition mean as
\begin{align*}
\hat\epsilon_{t,i}^{\star}
&=
\frac{x_{t,i}-\sqrt{\bar\alpha_t}\,x_{0,i}}
{\sqrt{1-\bar\alpha_t}},
&
\mu_{t,i}^{\star}
&=
a_t x_{t,i}+b_t\hat\epsilon_{t,i}^{\star}.
\end{align*}
Here, $\bar\alpha_t$ denotes the cumulative signal-retention coefficient of
the shared diffusion schedule.
The term ``reference-derived'' describes the source of the state rather than a
model prediction. A reference-derived state is obtained by forward-noising the
native coordinates as
$x_{t,i}=\sqrt{\bar\alpha_t}x_{0,i}
+\sqrt{1-\bar\alpha_t}\epsilon_{t,i}$. Consequently,
$\hat\epsilon_{t,i}^{\star}$ recovers the sampled forward noise and
$\mu_{t,i}^{\star}$ equals the mean of the native-conditioned posterior
$q(x_{t-1,i}\mid x_{t,i},x_{0,i})$ under the standard DDPM parameterization
\citep{ho2020ddpm}. A rollout state is instead produced by H-DiffAb's reverse
trajectory and need not be a forward-noised realization of the paired native
structure. On such a state, the same mapping defines a counterfactual,
schedule-consistent native reference rather than the posterior mean of the
rollout-generating process. Reference-derived states therefore serve as a
forward-process control for determining whether the observed correction
geometry is specific to model-visited states.

For each generated CDR residue $i\in\mathcal{G}$, let
\[
\delta_{t,i}
=
\mu_{T,t,i}^{\mathrm{pos}}-\mu_{H,t,i}^{\mathrm{pos}},
\qquad
d_{t,i}^{\star}
=
\mu_{t,i}^{\star}-\mu_{H,t,i}^{\mathrm{pos}}.
\]
Viewed from the H-DiffAb transition mean, $\delta_{t,i}$ is the displacement to
the Teacher prediction, whereas $d_{t,i}^{\star}$ is the remaining displacement
to the native-conditioned target. We measure their cosine alignment and the RMS
magnitude of the Teacher correction as
\begin{equation}
\label{eq:correction-geometry}
\mathrm{Align}_t
=
\frac{
\sum_{i\in\mathcal{G}}\delta_{t,i}^{\top}d_{t,i}^{\star}
}{
\sqrt{\sum_{i\in\mathcal{G}}\|\delta_{t,i}\|_2^2}
\sqrt{\sum_{i\in\mathcal{G}}\|d_{t,i}^{\star}\|_2^2}
},
\qquad
\mathrm{Mag}_t
=
\sqrt{
\frac{1}{|\mathcal{G}|}
\sum_{i\in\mathcal{G}}\|\delta_{t,i}\|_2^2
}.
\end{equation}
An alignment near one indicates that the Teacher changes H-DiffAb in the same
direction as the native-conditioned displacement; zero indicates an orthogonal
change, and negative values indicate a change away from that direction.
Magnitude reports the RMS transition-mean displacement per generated residue in
\AA{}. Alignment is directional and magnitude is scale-only; neither alone
establishes improvement, which is assessed separately by Teacher advantage in
Fig.~\ref{fig:on-policy-controls-mechanism}(c). For each complex--seed state,
alignment is computed after concatenating the generated-residue C$\alpha$
vectors; results are then averaged over seeds and macro-averaged across
complexes. Coordinates remain in the fixed complex frame, and the coordinate
scale is restored before reporting magnitude.

\begin{figure}[t]
\centering
\includegraphics[width=\linewidth]{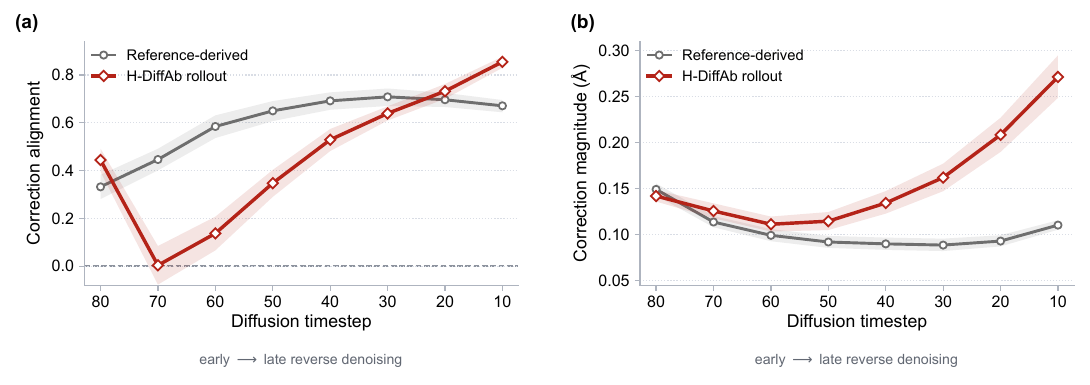}
\caption{\textbf{Teacher-correction geometry on RAbD CDR-H3.}
\textbf{(a)} Cosine alignment between the Teacher correction relative to
H-DiffAb and the native-directed correction; higher values indicate a more
native-directed correction.
\textbf{(b)} RMS correction magnitude in \AA{}, which reports correction size
rather than quality alone.
Gray curves show reference-derived (forward-noised native) states as a
forward-process control; red curves show H-DiffAb rollout states. Curves show
macro means over 60 complexes after averaging three fixed seeds; bands denote
95\% complex-level bootstrap confidence intervals.}
\label{fig:timestep-mechanism-diagnostics}
\end{figure}
Beginning at $t=50$, where Teacher advantage becomes positive,
$\mathrm{Align}_t$ on H-DiffAb rollout states increases from $0.347$ to $0.855$
at $t=10$, while $\mathrm{Mag}_t$ increases from $0.115$ to $0.272$~\AA{}.
The corresponding reference-derived curves remain comparatively stable over
the same interval (Fig.~\ref{fig:timestep-mechanism-diagnostics}). Thus, the
late-timestep Teacher advantage is accompanied by corrections that become both
larger and increasingly native-directed specifically on rollout states. This
provides a geometric interpretation of the stage-dependent transition behavior
rather than an independent measure of final generation quality.

\subsection{Per-CDR redesign}
\label{app:per-cdr-redesign}

\begin{table}[H]
\centering
\caption{\textbf{Per-CDR redesign on the 19 SAbDab test complexes used for simultaneous multi-CDR redesign.}
Each CDR is redesigned independently.
AAR is averaged over all target cases, whereas RMSD and ddG are reported separately for each redesigned CDR.
ABOPD achieves the highest overall AAR and the lowest H3 RMSD and ddG, while H-DiffAb performs better on several shorter or light-chain CDR metrics.
Best and second-best results are shown in \textbf{bold} and
\protect\underline{underlined}, respectively.}
\label{tab:single-cdr-region}
\scriptsize
\setlength{\tabcolsep}{2.5pt}
\begin{tabular}{lccccccccccccc}
\toprule
 & & \multicolumn{6}{c}{Per-CDR RMSD (\AA) $\downarrow$} & \multicolumn{6}{c}{Per-CDR ddG $\downarrow$} \\
\cmidrule(lr){3-8}\cmidrule(lr){9-14}
Method & AAR (\%) $\uparrow$ & H1 & H2 & \cellcolor{h3colblue}H3 & L1 & L2 & L3 & H1 & H2 & H3 & L1 & L2 & L3 \\
\midrule
DiffAb & 50.89 & 0.880 & 0.754 & \cellcolor{h3colblue}3.359 & 1.212 & 1.050 & 1.070 & 4.45 & 5.17 & 31.17 & 4.64 & 3.70 & \textbf{0.02} \\
H-DiffAb & \underline{61.72} & \underline{0.778} & \textbf{0.662} & \cellcolor{h3colblue}\underline{3.049} & \underline{0.954} & \textbf{0.991} & \textbf{0.987} & \underline{2.56} & \textbf{0.28} & \underline{20.31} & \underline{3.01} & \textbf{-0.20} & \underline{1.89} \\
ABOPD & \textbf{62.29} & \textbf{0.727} & \underline{0.688} & \cellcolor{h3colblue}\textbf{2.547} & \textbf{0.945} & \underline{1.032} & \underline{0.996} & \textbf{1.14} & \underline{0.59} & \textbf{11.52} & \textbf{2.86} & \underline{0.89} & 3.17 \\
\bottomrule
\end{tabular}
\end{table}

\subsection{EMA analysis}
\label{app:ema-analysis}
\begin{table}[H]
\centering
\caption{\textbf{EMA ablation across evaluation settings.}
The upper block compares ABOPD with and without EMA, while the lower reports paired per-CDR differences computed as EMA minus non-EMA; negative \(\Delta\)RMSD and positive \(\Delta\)AAR favor EMA.
EMA improves all aggregate metrics in the upper block and per-CDR RMSD for five of six CDRs, with a \(+0.56\) trade-off in aggregated per-CDR ddG.
Best and second-best results are \textbf{bold} and \protect\underline{underlined}, respectively.}
\label{tab:ema-ablation}
\small
\begin{tabular}{lccccc}
\toprule
 & \multicolumn{3}{c}{Simultaneous multi-CDR redesign} & \multicolumn{2}{c}{RAbD CDR-H3} \\
\cmidrule(lr){2-4}\cmidrule(lr){5-6}
Method & AAR (\%) $\uparrow$ & RMSD (\AA) $\downarrow$ & ddG $\downarrow$ & AAR (\%) $\uparrow$ & RMSD (\AA) $\downarrow$ \\
\midrule
ABOPD w/o EMA & \underline{60.53} & \underline{1.5357} & \underline{26.32} & \underline{36.25} & \underline{1.9954} \\
ABOPD w/ EMA & \textbf{60.73} & \textbf{1.4972} & \textbf{20.57} & \textbf{37.31} & \textbf{1.9500} \\
\bottomrule
\end{tabular}

\vspace{0.8em}
\setlength{\tabcolsep}{3.5pt}
\begin{tabular}{lcccccccc}
\toprule
\multicolumn{9}{l}{Paired effect of EMA on per-CDR redesign} \\
\midrule
& & \multicolumn{6}{c}{$\Delta$RMSD (\AA) $\downarrow$} & \\
\cmidrule(lr){3-8}
Difference & $\Delta$AAR (pp) $\uparrow$ & H1 & H2 & \cellcolor{h3colblue}H3 & L1 & L2 & L3 & $\Delta$ddG $\downarrow$ \\
\midrule
w/ EMA $-$ w/o EMA & +0.66 & -0.007 & -0.005 & \cellcolor{h3colblue}-0.055 & -0.033 & +0.002 & -0.003 & +0.56 \\
\bottomrule
\end{tabular}
\end{table}

\subsection{CDR-H3 interface and side-chain diagnostics}
\label{app:hcdr3-diagnostics}


We additionally report interface-improvement percentage (IMP) and side-chain
dihedral realism to distinguish backbone-geometry gains from thresholded
interface behavior and side-chain packing. IMP is the percentage of designs
whose Rosetta InterfaceAnalyzer energy improves over the native reference after
the common relaxation pipeline; JSD$_{\mathrm{sc}}$ is computed after
side-chain packing and relaxation, with lower values indicating closer
side-chain dihedral distributions. Table~\ref{tab:hcdr3-interface-sidechain-diagnostics}
reports these additional diagnostics.

\begin{table}[H]
\centering
\caption{CDR-H3 interface-improvement and side-chain-dihedral diagnostics. IMP is reported as a percentage; lower JSD$_{\mathrm{sc}}$ is better. Best results are \textbf{bold} and second-best are \protect\underline{underlined}.}
\label{tab:hcdr3-interface-sidechain-diagnostics}
\small
\begin{tabular}{lcc}
\toprule
Method & IMP (\%) $\uparrow$ & JSD$_{\mathrm{sc}}$ $\downarrow$ \\
\midrule
DiffAb & 15.84 & \textbf{0.2368} \\
H-DiffAb & \textbf{24.84} & \underline{0.2451} \\
ABOPD & \underline{23.32} & 0.2566 \\
\bottomrule
\end{tabular}
\end{table}


ABOPD attains the best CDR-H3 RMSD in Table~\ref{tab:rabd-hcdr3} and lowers
mean ddG in the simultaneous and per-CDR evaluations in
Tables~\ref{tab:all-cdr-joint} and~\ref{tab:single-cdr-region}, but its IMP
remains slightly below H-DiffAb and its JSD$_{\mathrm{sc}}$ is higher. Thus,
lower mean backbone RMSD and ddG do not imply improved side-chain realism or a
larger fraction of designs crossing the $\mathrm{ddG}<0$ threshold.

\subsection{Additional qualitative CDR-H3 cases}
\label{app:qualitative-cases}

\begin{figure}[H]
\centering
\includegraphics[width=0.9\linewidth]{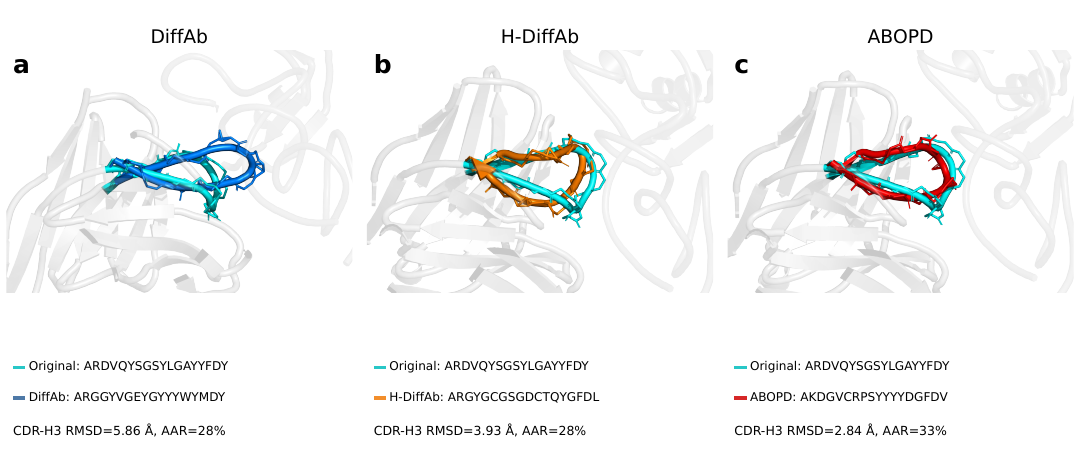}
\caption{Qualitative CDR-H3 comparison for RAbD target 4fqj.}
\label{fig:appendix-hcdr3-4fqj}
\end{figure}

\begin{figure}[H]
\centering
\includegraphics[width=0.9\linewidth]{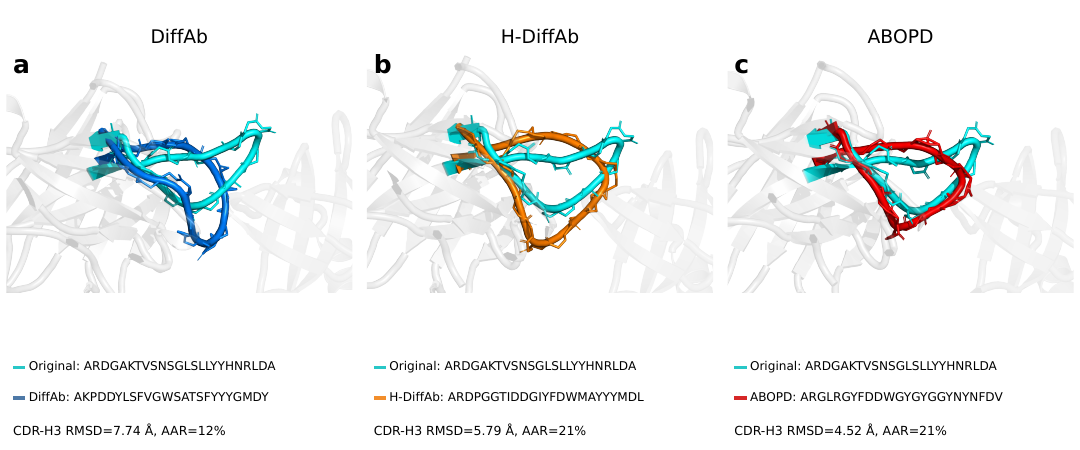}
\caption{Qualitative CDR-H3 comparison for RAbD target 4ot1.}
\label{fig:appendix-hcdr3-4ot1}
\end{figure}

\begin{figure}[H]
\centering
\includegraphics[width=0.9\linewidth]{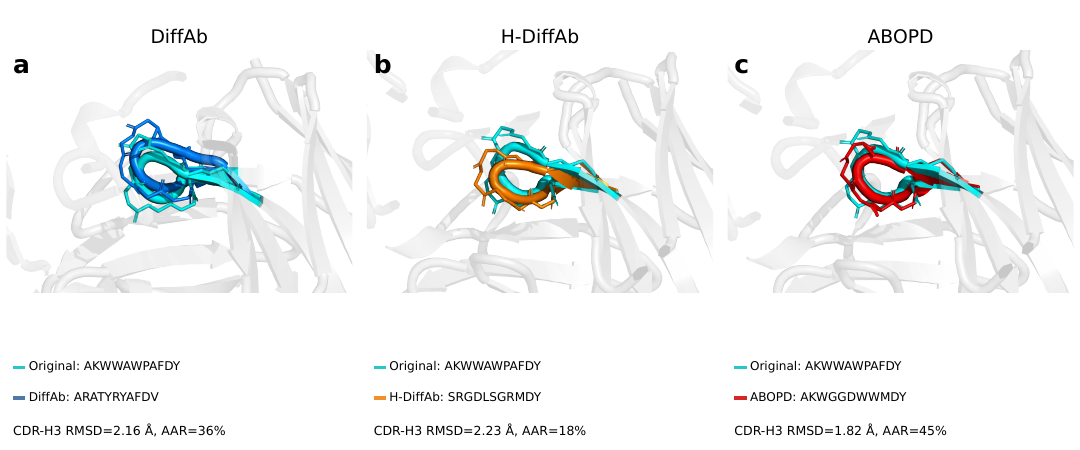}
\caption{Qualitative CDR-H3 comparison for RAbD target 3k2u.}
\label{fig:appendix-hcdr3-3k2u}
\end{figure}

\begin{figure}[H]
\centering
\includegraphics[width=0.9\linewidth]{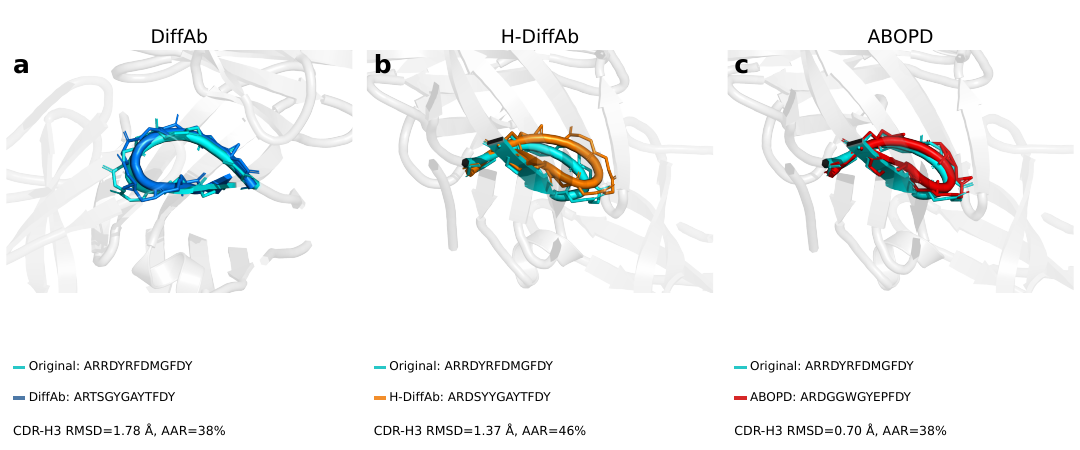}
\caption{Qualitative CDR-H3 comparison for RAbD target 5ggs.}
\label{fig:appendix-hcdr3-5ggs}
\end{figure}

\begin{figure}[H]
\centering
\includegraphics[width=0.9\linewidth]{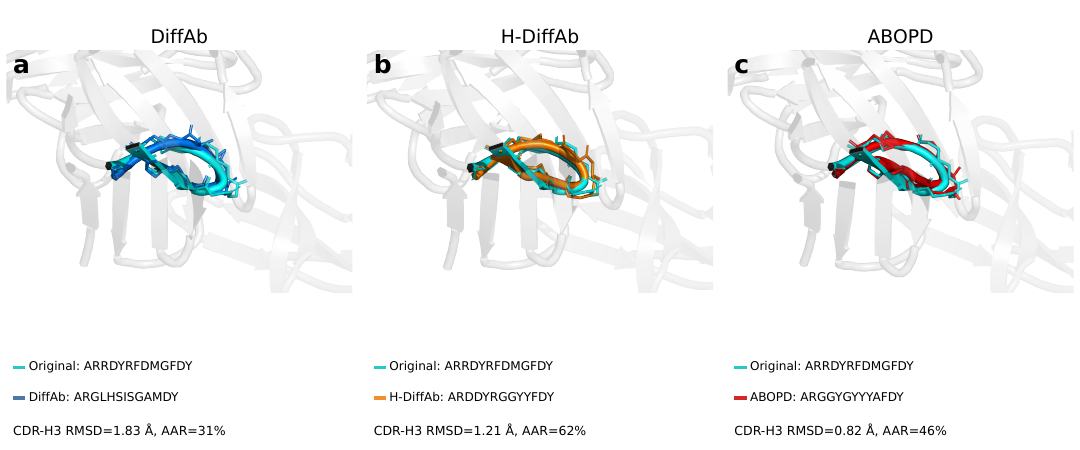}
\caption{Qualitative CDR-H3 comparison for RAbD target 5b8c.}
\label{fig:appendix-hcdr3-5b8c}
\end{figure}

Figures~\ref{fig:appendix-hcdr3-4fqj}--\ref{fig:appendix-hcdr3-5b8c} show five additional RAbD cases, where only CDR-H3 is redesigned. They span short and difficult long-loop conformations; ABOPD generally follows the native trace more closely, although the hardest case remains challenging.

\Needspace{8\baselineskip}
\section{Training and Implementation Details}
\label{app:training-details}

Unless otherwise stated, coordinate preprocessing and diffusion scheduling
follow the public DiffAb implementation \citep{luo2022diffab}. All stages use
Adam, unit weights for the three denoising terms, and the same target-mask
distribution. Table~\ref{tab:training-details} summarizes the stage-specific
configuration. For the post-training controls, SFT uses the denoising anchor's
timestep sampling, whereas offline distillation uses the same fixed
teacher--student matching timesteps as ABOPD; both are evaluated without EMA.
ABOPD uses its EMA student unless otherwise specified. H-DiffAb pretraining
processes approximately $8\times$ as many examples as the default DiffAb
configuration.

\begin{table}[H]
\centering
\scriptsize
\caption{Unified training configuration for H-DiffAb pretraining, backbone-aware teacher adaptation, and ABOPD post-training.}
\label{tab:training-details}
\setlength{\tabcolsep}{3.5pt}
\begin{adjustbox}{max width=\linewidth}
\begin{tabular}{>{\raggedright\arraybackslash}p{0.17\linewidth}>{\raggedright\arraybackslash}p{0.25\linewidth}>{\raggedright\arraybackslash}p{0.27\linewidth}>{\raggedright\arraybackslash}p{0.25\linewidth}}
\toprule
\textbf{Component} & \textbf{H-DiffAb pretraining} & \textbf{Teacher adaptation} & \textbf{ABOPD student} \\
\midrule
Target-mask distribution & \multicolumn{3}{c}{$p_{\mathrm{hyb}}(\texttt{single},\texttt{multi},\texttt{all})=(0.25,0.40,0.35)$} \\
Timestep sampling & $\mathcal{U}\{1,\ldots,100\}$ & $\mathcal{U}\{1,\ldots,100\}$ & Anchor: $\mathcal{U}\{1,\ldots,80\}$ \\
Rollout start index & -- & -- & $K=80$ on the original 100-step schedule \\
Rollout timesteps & -- & -- & $\{80,70,60,50,40,30,20,10,5\}$ \\
Distillation weights & -- & -- & Coordinate: $0.6$; sequence/orientation: $0$ \\
EMA usage & -- & -- & $\rho=0.9995$; updated after each student step; evaluation and sampling only \\
Learning rate & $10^{-4}$ & $10^{-4}$ & $10^{-5}$ \\
Training steps & 200k & 30k & 50k \\
Global batch size & 128 & 128 & 32 \\
\bottomrule
\end{tabular}
\end{adjustbox}
\end{table}

\phantomsection
\paragraph{Privileged descriptor and Teacher adaptation.}
\label{app:teacher-inputs}
The privileged descriptor $B$ is injected through two residual adapters
between the standard H-DiffAb encoder and EpsilonNet. It enriches the encoder
representations without replacing or modifying the noisy diffusion state
$R_t$; the student and Teacher therefore receive the same $R_t$, but only the
Teacher receives $B$.

For each target residue, the residue descriptor concatenates the native N,
C$\alpha$, C, O, and C$\beta$ coordinates, five corresponding atom-validity
masks, and a target-region indicator, yielding 21 input dimensions.
Coordinates are centered at the residue's native C$\alpha$ position and divided
by 10~\AA{}, while retaining the original Cartesian axes of the complex rather
than constructing a rotation-normalized residue frame. A
$21\!\rightarrow\!128\!\rightarrow\!128$ ReLU MLP maps this descriptor to a
masked residual that is added to the base residue embedding.

For each target--complex residue pair with C$\alpha$ distance $d_{ij}$, the pair descriptor contains
$\exp[-(d_{ij}/(10~\text{\AA}))^2]$,
$\min(d_{ij}/(100~\text{\AA}),1)$, and the target and context
C$\alpha$-validity indicators. A
$4\!\rightarrow\!64\!\rightarrow\!64$ ReLU MLP encodes the masked directed
features. Its output is averaged with its transpose and added to the base pair
embedding. Output masking ensures zero adapter contributions for non-target
residues and invalid pairs.

Explicit target amino-acid labels and atoms beyond C$\beta$ are excluded,
although the observed C$\beta$ coordinate and validity mask may carry limited
sequence-correlated information. In the main configuration, the H-DiffAb
backbone and both adapters are jointly optimized under the Teacher-adaptation
objective and frozen afterward. The two adapters add only 23.8K parameters in
total; adapter-only optimization is evaluated only as an ablation.
\section{Baseline and Evaluation Protocols}
\label{app:baseline-methods}

We distinguish locally reproduced baselines from published baseline results. In the simultaneous multi-CDR redesign benchmark (Table~\ref{tab:all-cdr-joint}), dyMEAN, AbX, PPIFlow, and IgGM were run by us on the same SAbDab split whenever an official implementation was available, using default hyperparameters unless a format bridge was required. DiffAb, H-DiffAb, and ABOPD are evaluated with the same sampling, relaxation, and diagnostic pipeline. H-DiffAb is our hybrid-pretrained DiffAb variant, and its parameters $\theta_0$ initialize SFT, offline distillation, and ABOPD. In the RAbD CDR-H3 benchmark (Table~\ref{tab:rabd-hcdr3}), the RosettaAb, MEAN, GeoAB, dyMEAN, DGENet, and BoltzGen rows are copied from \citet{wu2026proteor1} and \citet{kong2025unimomo}.

For a protocol-matched comparison with DiffAb, H-DiffAb, and ABOPD, we disabled PPIFlow's strict post-generation acceptance filter, which otherwise retains a sample only when the framework RMSD is below 1~\AA{} and no backbone clash is detected. In our SAbDab simultaneous multi-CDR run, this filter accepted only 64 of 1900 raw samples (3.37\%). We therefore evaluate all raw PPIFlow samples, so that PPIFlow is compared as a generator under the same sample budget rather than as a filtered candidate-selection pipeline.

\section{Limitations and Future Directions}
\label{app:limitations}

\paragraph{Structural fidelity and design diversity.}
ABOPD is designed to improve C$\alpha$ backbone trajectories, and its gains do
not extend uniformly to side-chain packing or interface improvement, as reflected
by JSD$_{\mathrm{sc}}$ and IMP in
Table~\ref{tab:hcdr3-interface-sidechain-diagnostics}. The lower RMSD and clash
rates, together with the more concentrated CDR-H3 sequence-space pattern in
Fig.~\ref{fig:generated-design-analysis}(a), are consistent with a
fidelity--diversity trade-off: dense trajectory supervision may concentrate
generation on structurally reliable modes while reducing exploration and limiting
gains in IMP. Future work could extend trajectory-level supervision from
C$\alpha$ backbone transitions to all-atom geometry, while developing
diversity-aware objectives or sampling strategies that better balance structural
fidelity with exploration of the sequence--structure space.

\paragraph{Task-aware trajectory supervision.}

ABOPD uniformly weights the selected rollout timesteps, whereas our
timestep-level transition analysis on shared H-DiffAb rollout states indicates
that the Teacher's local advantage and ABOPD's lower transition errors are
concentrated toward the end of reverse denoising
(Fig.~\ref{fig:on-policy-controls-mechanism}(c)--(d)). By contrast, on-policy
distillation for long-form mathematical reasoning exhibits larger token-level
reverse-KL losses in early output prefixes \citep{zhang2026prefixopd}.
Although autoregressive token positions and reverse-diffusion stages are not
directly equivalent, these differing patterns suggest that supervision
allocation may need to reflect task-specific generation dynamics. Because our
temporal evidence is limited to CDR-H3 generation with the current
DiffAb-based model, future work should examine adaptive timestep selection or
weighting across antibody CDR design, protein-binder design, and broader protein-design settings.

\end{document}